# Robust and Accurate Object Velocity Detection by Stereo Camera for Autonomous Driving

Toru Saito, Toshimi Okubo, Naoki Takahashi

*Abstract*— Although the number of camera-based sensors mounted on vehicles has recently increased dramatically, robust and accurate object velocity detection is difficult. Additionally, it is still common to use radar as a fusion system. We have developed a method to accurately detect the velocity of object using a camera, based on a large-scale dataset collected over 20 years by the automotive manufacturer, SUBARU. The proposed method consists of three methods: a High Dynamic Range (HDR) detection method that fuses multiple stereo disparity images, a fusion method that combines the results of monocular and stereo recognitions, and a new velocity calculation method. The evaluation was carried out using measurement devices and a test course that can quantitatively reproduce severe environment by mounting the developed stereo camera on an actual vehicle.

## I. INTRODUCTION

In recent years, the role of sensors for autonomous driving (AD) and advanced driver assistance system (ADAS) has increased. There are various types of sensors such as millimeter wave radar, lidar, and camera. Among them, the performance of camera has improved dramatically with the progress of image processing technologies such as deep learning. In addition, only cameras can detect patterns such as traffic signs and color of traffic lights. For these reasons, most recent vehicles are equipped with cameras as standard equipment. Furthermore, a stereo camera, one of the camera systems, is known as a versatile and cost-effective sensor. It has the features of camera and may also generate distance data like lidar.

However, the camera system has some constraints. Its detection performance is often restricted under environment such as night or rain. The accuracy of distance and velocity is lower than that measured by radars. For this reason, many vehicles are equipped with both cameras and millimeter wave radar as a fusion system. If the camera-only system can accurately and robustly detect objects, high-cost fusion systems will not be necessary.

In spite of this, most researches related to vision focus on only detection ratio, and very few focusing on accuracy and real-world robustness (Fig.1). Although the recent works of [1] and [2] present an approach of accurate detection of object distance using the stereo camera, they do not mention the robustness at severe environment such as rainy weather and the accurate detection of object velocity.

One reason for such limited studies is that, for many researchers, datasets are very limited. The famous CityScapes Dataset [3] focuses mainly on data at daytime good conditions.

All authors are with SUBARU Corporation {saito.toru, ookubo.toshimi , takahashi.naoki.b}@subaru.co.jp

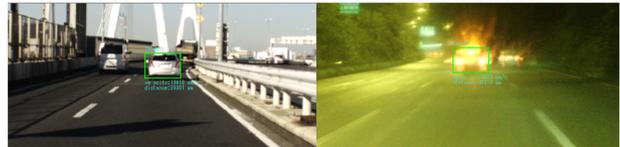

Fig.1 Detection results of our camera system. Left: Ideal environment for most research; Right: Bad environment in the real world. We focus on real world robustness at low cost using the camera-only system.

Although there are several studies by using datasets such as night [4] and fog [5], almost no data are found including more severe conditions such as dirt or raindrop on a windshield, etc., which are common environment for vehicle driving conditions. There are several reasons for this. It is difficult to create a common dataset suitable for researches, because the image differs depending on sensor types and exposure settings. Furthermore, the appearance of raindrops on the windshield greatly changes by the specifications of lens, the distance between the windshield and the camera, and the wiper system too.

In this work, we use a huge dataset in real world collected over 20 years by the automotive manufacturer, SUBARU [6], and propose some of key technologies to detect the velocity of object robustly and accurately.

First, we developed a new HDR detection approach using multiple stereo images with different exposures. When an image is captured with a bright exposure in a rainy environment at night, the light of the vehicle brake lamp is reflected in the raindrops, and it makes difficult to generate the accurate distance of the vehicle. Therefore, we use two distance images that are a stereo distance image generated with bright exposure and a stereo distance image generated with dark exposure, as well as the method to compare the reliability in post-processing to detect an object.

Second, we developed three fusion systems with functions of stereo-based object detection, mono-based object detection by right camera, and another mono-based object detection by left camera, and the method for switching those recognition outputs appropriately.

Finally, we developed a new velocity calculation method suitable for image recognition that can balance responsiveness and stability even in bad environment, instead of traditional Kalman filter [7].

We summarize our main cntributions as follows:

1.Proposal of multi-exposure stereo disparity fusion.

2.Proposal of stereo and mono detection fusion.

3.Proposal of new velocity calculation method suitable for image recognition in a real-world environment.

## II. RELATED WORKS

The devices for detecting the distance and velocity of objects using a stereo camera have been developed for a long time. Among them, [6] is known as the world's first mass-production vehicle with an in-vehicle stereo camera. In the past, the devices consisted of small VGA-sized CCD cameras with very limited resolution, brightness, frame rate, and dynamic range. However, the recent dramatical improvement of image sensors allows us to get a large amount of information.

What is important for a camera device is to improve the accuracy of an image obtained from a sensor. In a real-world road environment, it is necessary to acquire an image with a very high dynamic range, such as acquiring a dark object at night and a brake lamp of a vehicle at the same time. In order to solve such a problem, [8] proposes a method of generating a stereo image for a pair of stereo matching images captured with three different exposures and combining the stereo images. However, the Middlebury Stereo Datasets used in that research are ideal and far from the road environment we have to adapt.

Since the stereo camera has two monocular cameras, the distance and velocity can be calculated not only with the stereo camera but also with the monocular camera. There are several methods for calculating the distance and velocity using a monocular camera. Stein et al. [9] have proposed a distance estimation method based on the angle of view at the bottom of the preceding vehicle. However, this method will cause errors when pitching or on slopes. Although distance and velocity can be calculated from local features such as edges and corners on the image [10], it is difficult to calculate local features in bad weather. In addition, a method of calculating the distance and velocity by fusion between a monocular camera and a stereo camera has been proposed [11], but errors occur in both the stereo camera and the monocular camera when the image is disturbed in bad weather. We propose a method to calculate the reliability of a stereo camera and a monocular camera and to improve the stability in bad weather by considering the case where both are unreliable.

Finally, a very important technique for detecting an object with a camera is the method of velocity calculation. While there are many researches of techniques for detecting an object in an image, the methods for calculating the velocity of the object has not changed for a long time. For example, the Kalman filter proposed by Kalman in 1960 [8] is still being used in many fields. In general, the method to calculate the velocity from the image is calculated by differentiating the distance between the times, but the accuracy is poor at far distance and severe environment. The Kalman filter can calculate an appropriate value corresponding to the amount of noise in the observation, but the response tends to be delayed. Although the Kalman filter is a general-purpose calculation method for various observation targets and has been extended to a method for fitting more complicated models, this problem has not been improved basically. We propose a method to improve this problem by performing filtering focusing on the vehicle motion characteristics observed in the image.

## III. METHODOLOGY

In this chapter, we describe our method for performing robust and accurate object recognition with the stereo camera.

### A. The Architecture of Our Stereo Camera

The stereo camera we have developed consists of two CMOS image sensors at a frame rate of about 50ms (Fig.2). The CMOS sensors are capable of outputting multiple exposures (T1 and T2), and T1 exposure and T2 exposure are sequentially captured. The distance of the object is calculated from the disparities from the stereo camera using the trigonometric function (1). Baseline length $L$, focal length $f$, and sensor pitch $w$ are the fixed parameters in the system, and in our research, the camera specifications are written by (2).

The stereo disparity is calculated by the FPGA unit, and the image processing of object detection is performed by the CPU. In addition, in order to accelerate legacy machine learning that can be performed with very compact calculation, the integrated images of brightness and the edge strength are calculated by the FPGA.

$$D = \frac{L \times f}{w \times d} \qquad (1)$$

$$D = 560000 / d \qquad (2)$$

$D$=Distance(mm)
$d$=disparity(pix)

### B. Multi-Exposure Disparity Fusion

The developed stereo camera captures two exposure images, long exposure for object and lane detection (T1) and short exposure for light and traffic sign detection (T2), in one frame (Fig.3). High dynamic range can be obtained with a single image by HDR rendering of a multi-exposure image. However, in order to accurately and robustly detect various objects even at night, it is necessary to cover 1000 times the range. When such a wide range is combined, information loss at the time of compression is large and accurate recognition cannot be performed. For this reason, we use two exposure images specific to the target without compositing and compressing.

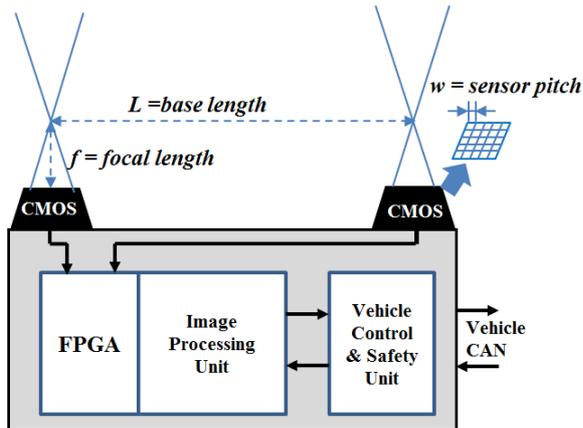

Fig. 2 The architecture of our stereo camera system.

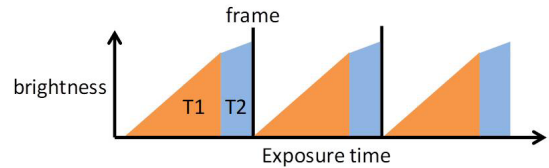

Fig.3 Our multi-exposure.T1 is long exposure for object and lane detection. T2 is short exposure for light and traffic sign detection.T1 and T2 are captured in one frame.

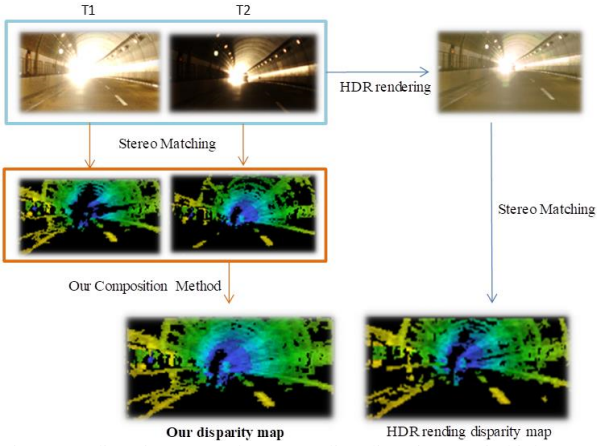

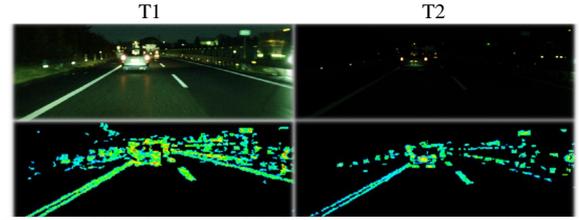

Fig.5 Reliability heat map. T1 disparity map is more reliable and adapted in dark scene.

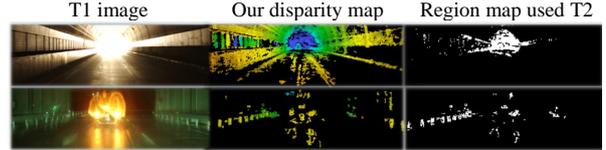

Fig.6 Region map used T2 disparity. T2 disparity contributes in bad environment.

Fig.4 Our disparity map and HDR rending disparity map in high dynamic range scene. Ours is more dense without noise.

Also, if the stereo matching is performed after the HDR synthesis, the disparity accuracy deteriorates due to the non-linear compression. Therefore, the stereo matching is performed on each of the two exposure images. However, since non-compression synthesis is possible between disparity maps, each disparity map is synthesized based on our composition method to create a disparity fusion map. Thereby, a single disparity map having a high dynamic range can be used without lowering the disparity accuracy.

*Our Composition Method*

We use a method of performing stereo matching processing without compression on each of T1 and T2, and combining T1 disparity map and T2 disparity map later. At the corresponding coordinates of the T1 disparity map and the T2 disparity map, more reliable disparity is adopted (Fig.4).

The stereo matching method outputs a sparse disparity map that consists of only reliable disparities. At the same time, the degree of reliability calculated from the edge strength is added as shown in reference [6]. At the coordinates of interest, it is determined which of the disparities T1 and T2 to use based on the reliability shown in the TABLE I. Basically, it fills in the form, but if both T1 and T2 have disparity, the larger of reliability is adopted. If reliability is also the same, T2 is preferentially adopted because dark images are less likely to produce left-right differences due to disturbances such as raindrops near the light source (Fig.5) (Fig.6). For this reason, there are still some topics to research in order to consistently adopt T2 disparity.

*C. Stereo and Mono Detection Fusion*

Second, we describe the method of fusion of the results of detections from mono and stereo images. When using a stereo image, if there are raindrops or dirt, it is difficult to calculate the accurate velocity of the preceding vehicle. In our system, the left and right monocular cameras detect the preceding vehicle and select the camera with better conditions for the detection (Fig.7). Then, the velocity is calculated by the monocular camera or stereo camera. In the velocity calculation using the monocular camera, a method was used

TABLE I. T1, T2 disparity composition judgment table

|  |  | T1 |  |
|---|---|---|---|
|  |  | *Disparity exist* | *Disparity none* |
| **T2** | *disparity exist* | if T1 Reliability > T2 Reliability then<br>  adopt T1<br>else:<br>  adopt T2 | adopt T2 |
|  | *disparity none* | adopt T1 | None |

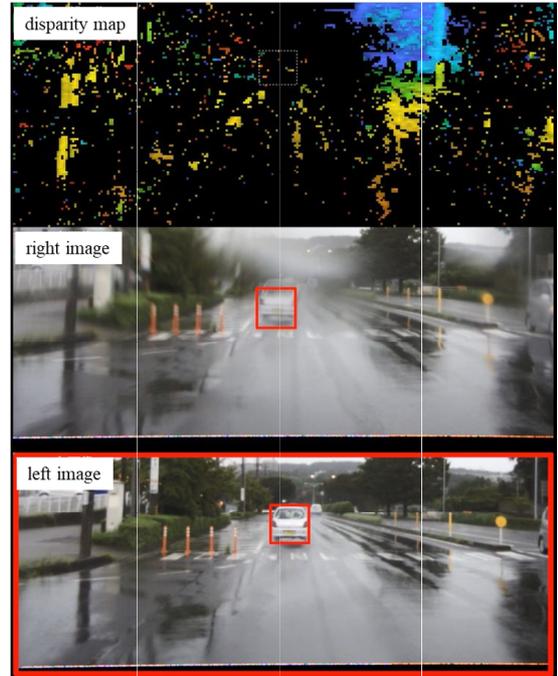

Fig.7 Top: Stereo disparity map. Middle: Right camera image. Bottom: Left camera image. In this scene, the result of detection from the left camera is selected.

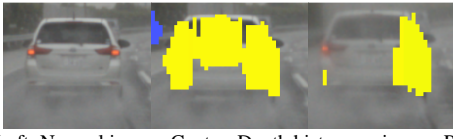

Fig.8 Left: Normal image. Center: Depth histogram image. Right: Depth histogram image in bad weather.

based on changes in the width of the image of the preceding vehicle.

In bad weather, the velocity may be unstable even when the distance can be calculated by the stereo camera. Therefore, our system uses the stereo velocity ($V_s$), monocular velocity ($V_m$), and predicted velocity ($V_p$) for stabilization and outputs the fusion velocity $V_f$ as:

$$V_f = \begin{cases} \dfrac{V_s \times W_s + V_m \times W_m + V_p \times W_p}{W_s + W_m + W_p}, (R_s \leq 2) \\ V_s, (R_s > 2) \end{cases} \quad (3)$$

$V_s$ is the velocity of the stereo and uses the difference from the past stereo distance to calculate the velocity. $R_s$ is the stereo reliability and is calculated based on the depth histogram of the preceding vehicle. The depth histogram is a histogram that summarizes the same depth distance of the height direction. In bad weather, the number of depth histograms also decreases because the distance that can be calculated decreases. In such case, the calculation of distance and velocity becomes unstable, so the stereo reliability is calculated based on the number of depth histograms (Fig.8).

The threshold for the number of depth histograms used for reliability depends on the distance and is determined by the four states; TRUST, STABLE, MAYBE, and NONE (Fig. 9). Depending on the determined state, the stereo reliability $R_s$ and stereo weight $W_s$ are defined as shown in Table II. The monocular weight $W_m$ is defined as follows.

$$W_m = (3 \times R_m)^2 \quad (4)$$

$R_m$ is the monocular reliability in the range of 0 to 1. This value is calculated based on the similarity between the texture (edge histogram) and the previous image, the amount of change in the angle of view at the bottom of the preceding vehicle, and so on. $V_p$ is the predicted velocity and is calculated as follows.

$$V_p = \begin{cases} \min(V_{t-1}, V_{t-S} + A_{t-S} \times S \times \Delta t), (A_{t-S} < -0.1G) \\ V_{t-1}, (else) \end{cases} \quad (5)$$

$A_{t-S}$ is the acceleration of the preceding vehicle before $S$ frame, that is the latest frame when the stereo distance could be calculated. $V_{t-1}$ and $V_{t-S}$ represent the velocity 1 frame before and $S$ frame before. $\Delta t$ is the length of 1 frame time If both the stereo velocity and the monocular velocity become unstable when the preceding vehicle decelerates to less than -0.1G, it is dangerous if the preceding vehicle approaches, so the velocity in consideration of deceleration is predicted. The expected weight $W_p$ is defined as follows.

$$W_p = (3 - R_s) \times (3 - 3 \times R_m) \quad (6)$$

As described above, both stability and responsiveness to deceleration from a preceding vehicle are achieved even in bad weather.

*D. Velocity Filter*

Then, we describe the method of calculating the velocity of an object. This is basically written in a very simple formula as shown in (7). The velocity is calculated by differentiating the object distance $D_t$ detected from the latest image captured, the object distance $D_{t-1}$ detected from the image captured at 1 frame before, at the imaging time interval $\Delta t$.

$$V_t = \dfrac{D_t - D_{t-1}}{\Delta t} \quad (7)$$

V=Velocity(mm/s)
D=Distance(mm)
t=time(s)

In this case, filtering process is performed generally because the raw velocity has large errors. The typical method of the filtering process is to compare the previous filtered velocity $VN_{t-1}$ with the current raw velocity $V_t$ and use the weight $GV$ to determine how much it affects the final filtering velocity $VN_t$ (8). Our proposed filter basically has the same structure, but the feature is to use $Gain=S$ which takes into account the past motion of the object (9). The detail of the method for determining $S$ is described as follows.

$$VN_t = VN_{t-1} + GV \times (V_t - VN_{t-1}) \quad (8)$$

TABLE II. STEREO RELIABILITY $R_s$ AND STEREO WEIGHT $W_s$

| | $R_s$ | $W_s$ |
|---|---|---|
| TRUST | 3 | 9 |
| STABLE | 2 | 4 |
| MAYBE | 1 | 1 |
| NONE | 0 | 0 |

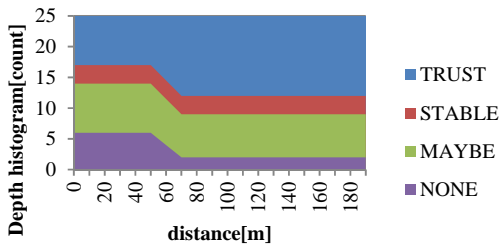

Fig.9 Relationship of depth histogram and distance.

$$VS_t = VS_{t-1} + S \times (V_t - VS_{t-1}) \quad (9)$$

*GV = gain for velocity* [min:0, max:1]
*VN (mm/s) = filtered velocity by normal method*
*VS (mm/s) = filtered velocity by Saito method (ours)*
*S = Saito Gain*

*S* is basically determined based on the acceleration of the object (10). $AN_{t-1}$ is the filtered acceleration at 1 frame before. It can be calculated by the normal filtering method in every frame (11). The feature of our filter is that when comparing this $AN_{t-1}$ with the current raw acceleration $AS_t$, $AN_{t-1}$ is multiplied by *B(=Bias Gain)* (10). The raw acceleration $AS_t$ is calculated by differentiating the difference between the raw velocity $V_t$ and the previous velocity $VS_{t-1}$ (12), but the value generally includes very large errors. If the change between the previous acceleration and the current acceleration is large, normally the weight should be large. But in our method, if an error occurs in the same direction as $AN_{t-1}$, a strong weight is used, and if an error on the opposite side occurs, a weak weight is used, by using the value of the previous acceleration $AN_{t-1}$ amplified by Bias (10). The output *S* is divided from the value of *N* to normalize the gain so that it is used with a weight of *0* to *1*, and limit the weight *S* so that it does not become too strong (10) (13).

$$S = \frac{N}{|(AN_{t-1} \times B) - AS_t|} \quad (10)$$

$$AN = AN_{t-1} + GA \times (\frac{VS_t - VS_{t-1}}{\Delta t} - AN_{t-1}) \quad (11)$$

$$AS_t = \frac{V_t - VS_{t-1}}{\Delta t} \quad (12)$$

$$S = \min(LTh, S) \quad (13)$$

*N=Normalize Parameter*
*B=Bias Gain Parameter*
*AS (mm/s^2) = Acceleration*
*AN (mm/s^2) = filtered Acceleration by normal method*
*GA=Gain for Acceleration ([min:0, max:1])*
*LTh=Limit Threshold*

However, once the ground truth of the velocity and the calculated velocity *VS* greatly deviates, the value $AS_t$ always becomes large, and it may remain diverged without being converged for a long time. In order to solve this problem, the system is extended by adding a process of monitoring whether the value is greatly out of range.

The monitoring process is performed by *SM (S for Monitor)*, which is similar to *S* (14). *SM* is calculated by comparing the value obtained by applying the bias *B* to the previous acceleration and the current acceleration same as *S*. The acceleration $AM_t$ is calculated not by the velocity $VS_t$ but by the velocity $VN_t$ by the normal filtering method (15).

$$SM = \frac{N}{|(AN_{t-1} \times B) - AM_t|} \quad (14)$$

$$AM_t = \frac{V_t - VN_{t-1}}{\Delta t} \quad (15)$$

*SM=Saito Gain Monitor*
*AM(mm/s^2)=Acceleration for Monitor*

Finally, we describe a method for rejecting *S* by *SM* (Algorithm I). First, it is determined whether the value of *S* is abnormally small by comparing it with *MTh*. Then, in order to determine whether *S* is sufficiently smaller than *SM*, for example, the value is compared with a value obtained by multiplying *SM* by *RT*, and if the threshold is satisfied, *S* is rejected and the value of *SM* is applied. However, in order to prevent the value of *SM* from affecting too much, the value of *LThM* is limited so that the value of *SM* does not become too large.

## IV. EXPERIMENT

### A. Experimental Setup and Data.

In our experiments, we mounted a stereo camera on an actual vehicle and measured the velocity in an actual driving environment. There are mainly two types of experiments. One is the test driving confirmation by trained test drivers on data of over 400,000 km on actual roads around the world. Second is more quantitative measurement, in which two types of opposing characteristics, responsiveness and stability are measured with respect to a specified operation pattern of a preceding vehicle on a test course. This paper describes the latter quantitative evaluation results.

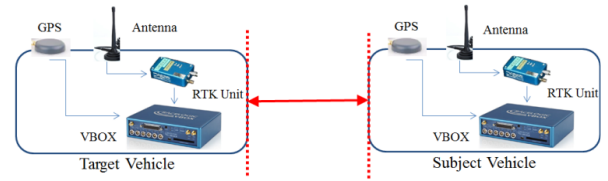

Fig.10 VBOX3iSL RTK GPS Unit

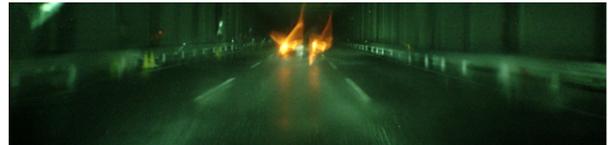

Fig.11 Image under test condition (30mm/h artificial rainfall)

**Algorithm I: Limit**

**if** S < MTh **then**
  **if** S < (SM×RT) **then**
    S = min(LThM,SM)
  **end if**
**end if**

*MTh=Monitor Threshold parameter*
*RT = Reject threshold parameter*
*LThM = Limit threshold parameter*

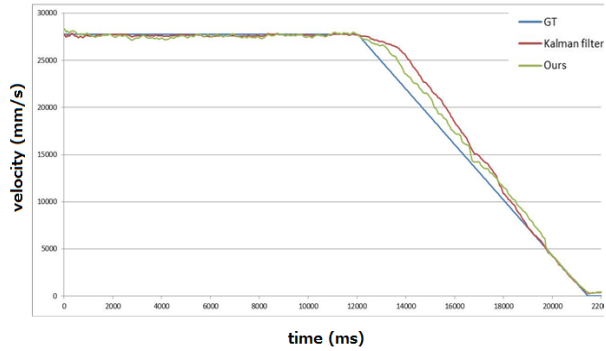

Fig.12 Graphs show the velocity of the preceding vehicle in case of deceleration pattern of 100 kph to 0 kph 0.3 G.

TABLE III.  THE DELAY FOR GT AT 72KPH IN FIG.12.

| Filter | Delay |
| --- | --- |
| Kalman Filter | 970ms |
| Ours | 485ms |

### A1. Measurement Device

Data measurement includes the true value of the moving velocity of the object. The two cars are equipped with a VBOX data logger that can log distance/velocity with the ± 27.8mm/s accuracy GPS (Fig.10). The velocity calculated by this is used as a ground truth. Vehicle movements are tested by several patterns of velocity, distance, and movement, from constant to deceleration, from constant to acceleration, from acceleration to deceleration, etc.

### A2. Test Course

In addition to measuring the characteristics in good environment, the characteristics in bad environment, which had been difficult to quantify, are also measured. The test was conducted at JARI's (Japan Automobile Research Institute) specific environment area. It is possible to mechanically generate a certain amount of precipitation in this course. The tests were conducted based on rainfall of 30 mm/h (Fig.11).

### B. Algorithm Parameters

The parameters of velocity filter used in all our evaluations are shown in TABLE V.

### C. Results

Firstly, we describe the results of tests on the responsiveness and stability of the velocity using only the velocity filter and comparison results in a bad environment in which the disparity fusion and the mono fusion are added.

TABLE V.  VELOCITY FILTER PARAMETERS

| Parameters | Values |
| --- | --- |
| GV | 1 / (Distance(mm) / 3500 + 1) |
| N | 980 |
| B | 16 |
| GA | 1/21 |
| LTh | 1/5 |
| MTh | 1/17 |
| RT | 1/4 |
| LThM | 1/15 |

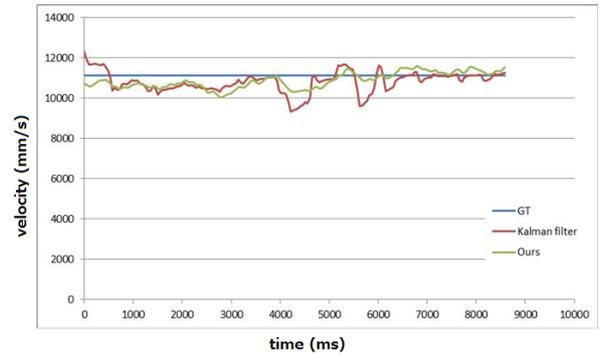

Fig.13 Graphs show the velocity of the preceding vehicle in case of 40 kph constant driving in raining condition.

TABLE IV.  THE DESPIRSION OF VELOCITY IN FIG.13.

| Filter | Dispersion |
| --- | --- |
| Kalman Filter | 466ms/s(1σ) |
| Ours | 392ms/s(1σ) |

Fig.12 shows the velocity responsiveness of our method compared to the traditional Kalman filter in a simple good environment. It is test data for a deceleration pattern (0.3 G) from a distance (about 55 m) where the conditions for calculating the responsiveness are severe. It has been confirmed that the response is higher than that of a traditional Kalman filter. The TABLE III shows the delay for GT at 72kph from 100kph.

On the other hand, Fig.13 shows the comparison results of stability in raining condition using the same parameters. While achieving high responsiveness, it is possible to output a relatively stable value even in the case of an input containing a large number of errors by the velocity filter.

Secondly, we show the comparison results in a bad environment, in which all of the three methods are combined, i.e. the multi-exposure disparity fusion, the stereo and mono detection fusion, and the velocity filter. Fig.14 shows the same deceleration pattern as in Fig.12 in raining condition. The graphs show that the combination of each method contributes to the final results. Comparing the target non-detection rate per frame of this scene, 12.23% without 1 and 2, 7.69% with 1 only, 0.00% with 1 and 2 . In such a non-constant speed scene, a high observation rate has an effect of responding without bias to prediction.

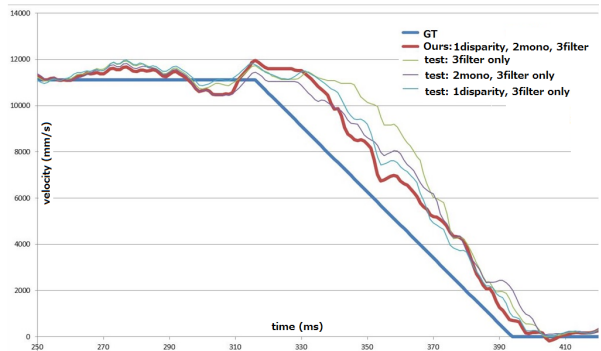

Fig.14 100 kph to 0 kph 0.3G deceleration in bad environment. 1: *Multi-exposure disparity fusion*, 2: *Stereo and mono detection fusion*, 3: *Velocity filter*. Graphs show the effectiveness.

V. Conclusion

In this work, we proposed that the robust and accurate object velocity detection can be achieved with the camera-only system. The use of cameras is expanding from ADAS to more advanced AD. We described the result of a simple scene when detecting the back of a preceding vehicle, mainly considering the application in ACC (Adaptive Cruise Control). In the future, however, a technology that detects road environments with more complex trajectories of objects will be needed. As we proposed, the detection method is important, and research for more complex objects needs to be developed in the future.